%% file: main.tex
\documentclass[letterpaper, 10 pt, conference]{ieeeconf}  

\IEEEoverridecommandlockouts                              

\usepackage{amsmath,amssymb}
\usepackage[ruled,linesnumbered]{algorithm2e}
\usepackage{algorithmic}
\usepackage{mathrsfs}
\usepackage{color}
\usepackage{cite}
\usepackage{multirow}
\usepackage{url}
\usepackage{xcolor}
\usepackage{xspace}
\usepackage{subfigure} 
\usepackage{graphicx}
\usepackage{comment}
\usepackage{threeparttable}

\newcommand{\dyfnote}[1]{ \color{blue} #1  \color{black}}

\makeatletter
\let\NAT@parse\undefined
\makeatother
\usepackage[colorlinks,
            urlcolor=blue,
            linkcolor=blue,       
            anchorcolor=blue,  
            citecolor=green        
            ]{hyperref}

\newcommand{\metric}{{p-Index}\xspace}
\newcommand{\core}{\emph{PFilter}\xspace}

\title{\LARGE \bf
\core: Building Persistent Maps through Feature Filtering for Fast and Accurate LiDAR-based SLAM
}

\author{Yifan Duan, Jie Peng, Yu Zhang, Jianmin Ji* and Yanyong Zhang
\thanks{* The corresponding author.}
\thanks{School of Computer Science and Technology, University of Science and Technology of China, Hefei, 230026, China
{\tt\small \{dyf0202, pengjieb\}@mail.ustc.edu.cn, \{yuzhang, jianmin, yanyongz\}@ustc.edu.cn}.}%
}%

\begin{document}

\maketitle
\thispagestyle{empty}
\pagestyle{empty}

\begin{abstract}
Simultaneous localization and mapping (SLAM) based on laser sensors has been widely adopted by mobile robots and autonomous vehicles. 
These SLAM systems are required to support accurate localization  with limited computational resources.
In particular, point cloud registration, i.e., the process of matching and aligning multiple LiDAR scans collected at multiple locations in a global coordinate framework, has been deemed as the bottleneck step in SLAM.
In this paper, we propose a feature filtering algorithm, \core, that can filter out invalid features and can thus greatly alleviate this bottleneck. Meanwhile, the overall registration accuracy is also improved due to the carefully curated feature points.  

We integrate \core into the well-established scan-to-map LiDAR odometry framework, F-LOAM, and evaluate its performance on the KITTI dataset. The  experimental  results  show  that  \core can remove about  48.4\%  of  the  points  in  the  local  feature  map  and  reduce feature points in scan by 19.3\% on average, which save 20.9\% processing time per frame. In the mean time, we improve the accuracy by 9.4\%. 

\end{abstract}

\input{chapters/introduction}
\input{chapters/relatedworks}
\input{chapters/preliminaries}

\input{chapters/method}

\input{chapters/experiment}

\input{chapters/conclusion}
\section*{ACKNOWLEDGMENT}

The work is partially supported by the 2030 National Key AI Program of China 2018AAA0100500, Guangdong Province R$\&$D Program 2020B0909050001, Anhui Province Development and Reform Commission 2020 New Energy Vehicle Industry Innovation Development Project and 2021 New Energy and Intelligent Connected Vehicle Innovation Project, Shenzhen Yijiahe Technology R$\&$D Co., Ltd..



\bibliographystyle{IEEEtran}
\bibliography{IEEEabrv, references}

\end{document}

%% file: chapters/introduction.tex
\section{Introduction}

Simultaneous localization and mapping (SLAM) plays a vital role in navigation of autonomous systems like autonomous carscite and robots~\cite{survey}. In recent years, with the development of LiDAR technology~\cite{survey_LiDAR}, LiDAR-based SLAM has attracted a great deal of attention because of LiDAR's capabilities of remaining robust against illumination changes and  obtaining accurate distance information~\cite{survey_LiDAR_slam}. 
In fact, LiDAR-based SLAM is often used to estimate the robot's ego-motion, denoted as LiDAR odometry (LO), and locate itself in the  map, denoted as localization~\cite{LOL}.
In this paper, we consider how to improve the efficiency and accuracy  of LO through better feature selection.


Point cloud registration, i.e., the  process  of matching and aligning multiple  LiDAR  scans collected at multiple locations in a global coordinate framework,
has been deemed as  a major time-consuming task in LO.
It is usually modeled as a nonlinear least-squares problem on suitable feature points, and solved by optimization methods like Gauss-Newton or Levenberg-Marquardt. Due to the large number of points in a LiDAR frame -- for example, a 64-line LiDAR has about 120 thousand points per frame -- typical registration methods such as ICP~\cite{icp} and NDT~\cite{ndt} usually involve a large amount of computations (e.g., nearest neighbor queries, gradient descent-based optimizations, etc.) with a risk of non-convergence~\cite{survey_registration}. {The other disadvantage stems from their sensitivity to initial poses, which can severely undermine the subsequent correspondence establishment process.}


To address the two issues, a popular approach~\cite{zhang2014loam} is to extract features from the raw data according to the geometric properties, and only use a portion of the point cloud for registration. Even with such downsampling operations, the point cloud registration still remains the bottleneck for many state-of-the-art SLAM systems~\cite{floam,pan2021mulls}, severely limiting the efficiency of the whole SLAM system.

Several schemes have been proposed to address this bottleneck through clever feature selection. For example, the schemes discussed in~\cite{greedy,li2021kfs,huangguoquan,chenzonghai} try to  reduce the number of constraints for registration by adding new geometric conditions or by evaluating the contribution or uncertainty of constraints to the optimization equation. 
While these schemes have explored the distribution of feature points in each individual point cloud frame, the distribution of feature points across consecutive frames are largely unexplored. 


\begin{figure}[t]
	\centering
	\subfigure[Local edge map before \core] {\includegraphics[width = 0.48\linewidth]{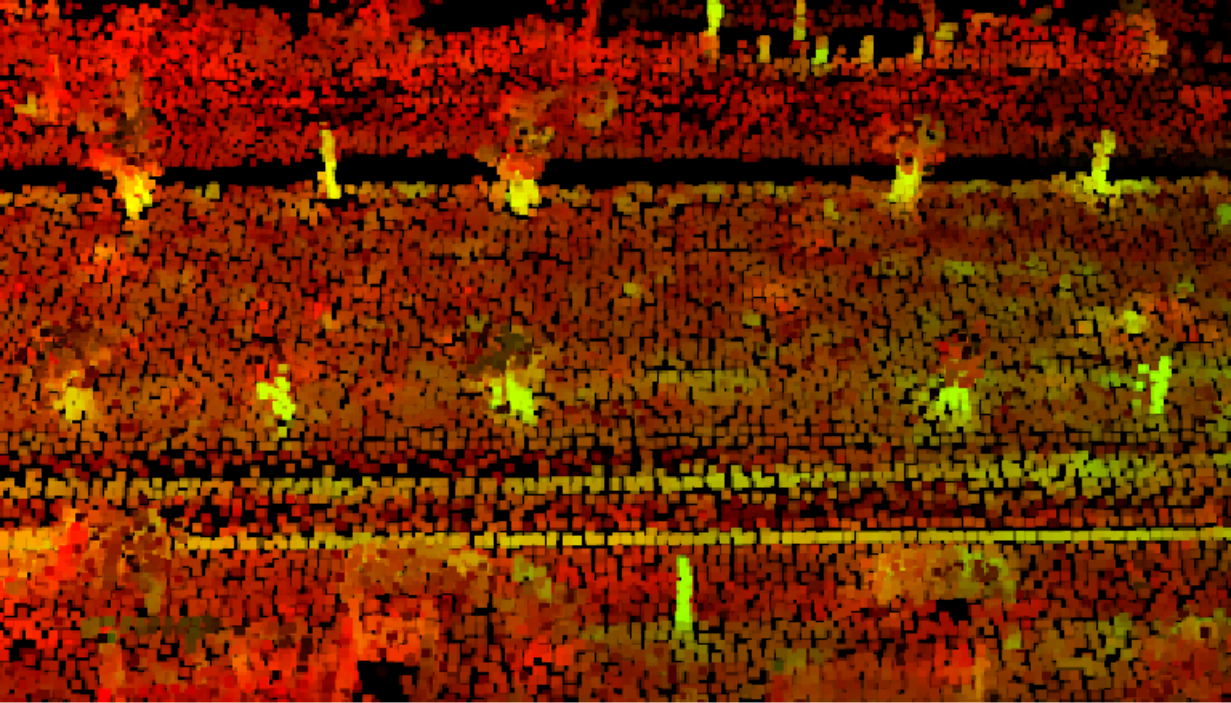}}
	\subfigure[Local edge map after \core] {\includegraphics[width = 0.48\linewidth]{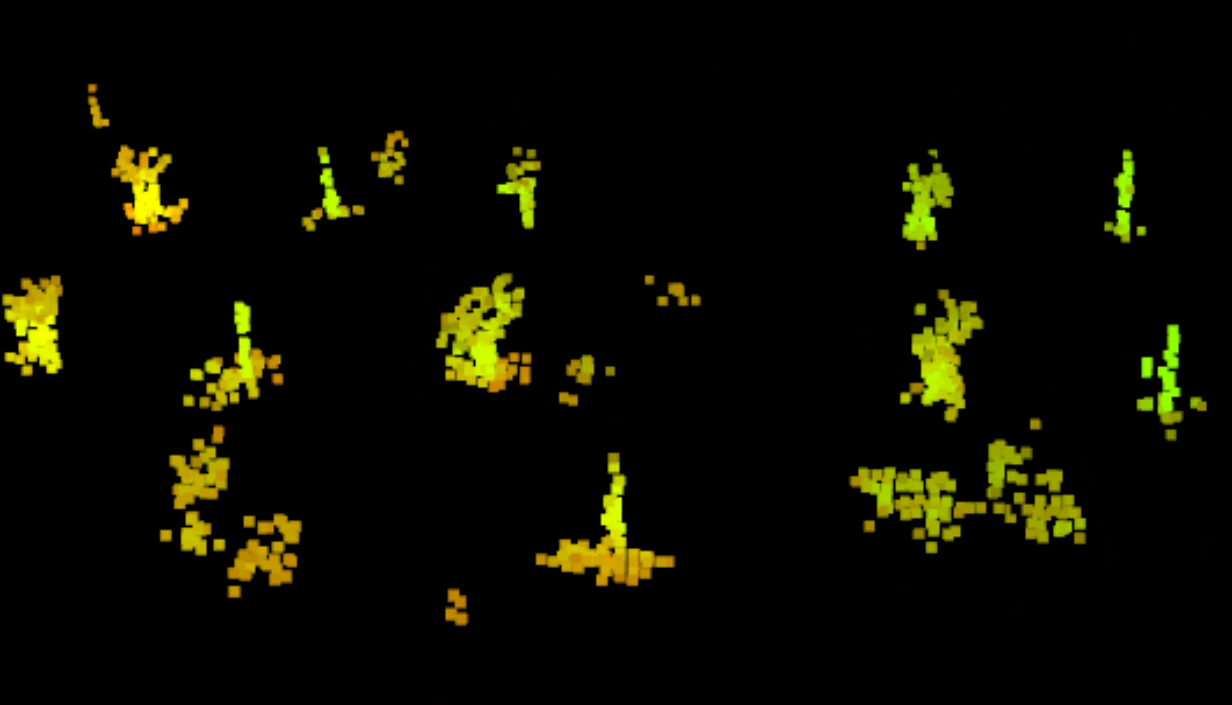}}
	\subfigure[Local surface map before \core] {\includegraphics[width = 0.48\linewidth]{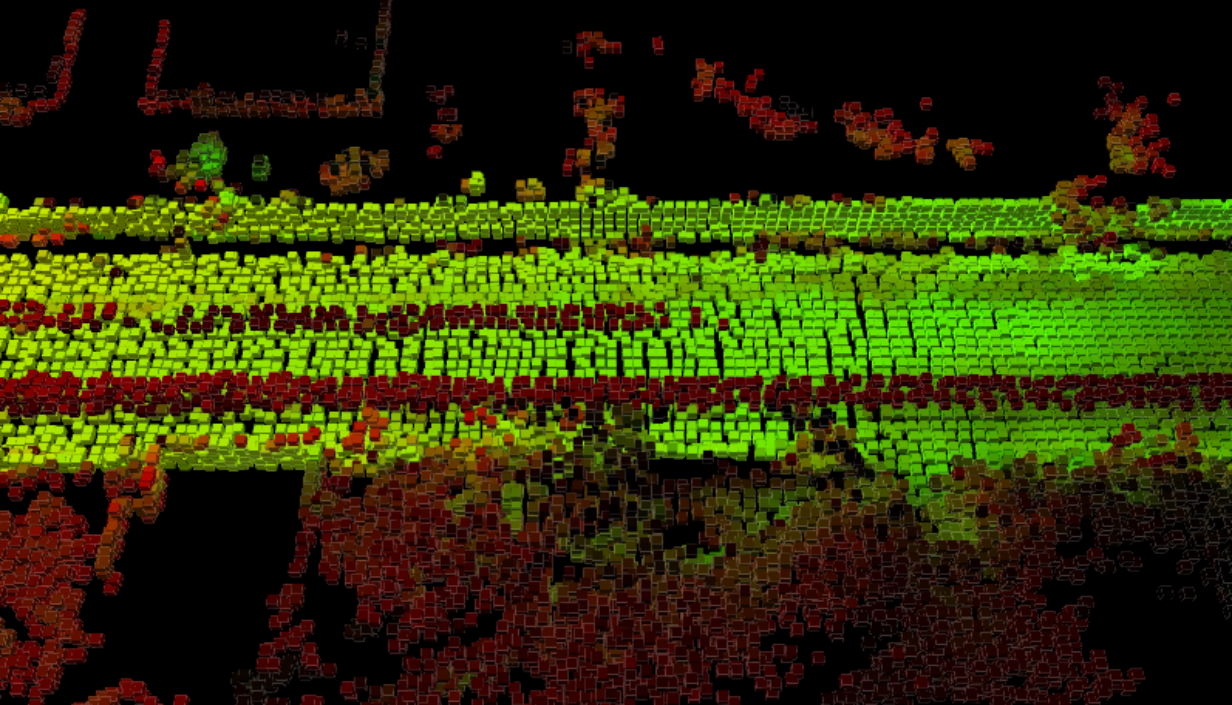}}
	\subfigure[Local surface map after \core] {\includegraphics[width = 0.48\linewidth]{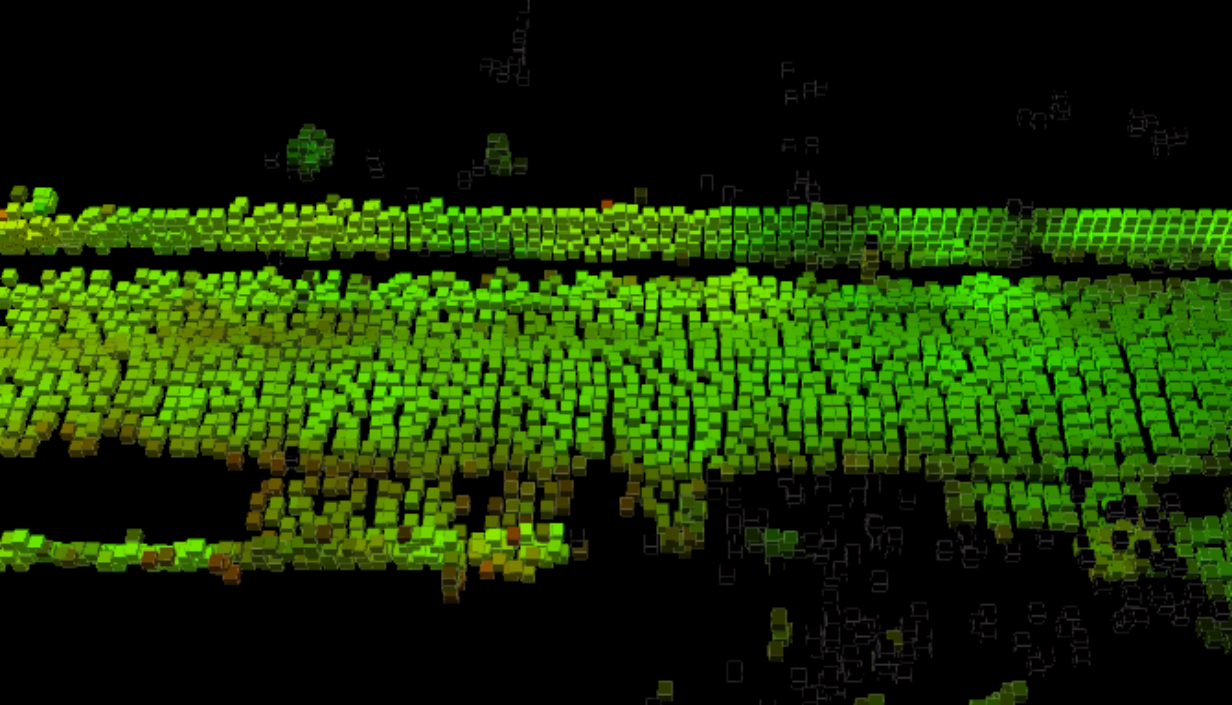}}
	\caption{The effect of \core for a local feature map consisting of a local edge map and a local surface map.
	Each point in the maps is colored by its \metric value: the red color indicates points with low \metric values, i.e., transient feature points, while the green color indicates points with high \metric values, i.e.,  persistent feature points. The local maps (a) and (c) contain a large number of incorrectly extracted features; after filtering out these erroneous feature points, the resulting maps in (b) and (d) contain fewer feature points that should be included for registration. Here, the edge map in (b) only keeps landmarks such as poles, trunks, while the surface map in (d) only keeps landmarks such as  ground and walls.}
	\label{fig:show}
	\vspace{-2.0em}
\end{figure}

In this paper, we set out to explore the cross-frame feature point distribution for more efficient feature selection. Towards this end, we propose a new attribute for feature points, named persistence-index, \metric in short. \metric evaluates whether a feature point is \emph{persistent} or \emph{transient} by observing {its} distribution 
across consecutive frames, which is weighed by how frequent a feature point is detected recently.
Intuitively, larger \metric values are assigned to those feature points that have been detected frequently in the near past.
We call feature points with large values of \metric as persistent. 

In our experiments, we find out that, feature points that are persistent usually correspond to stationary landmarks, like lamp poles, tree trunks, grounds, walls, etc. 
On the contrary, transient feature points are usually generated occasionally, possibly due to events such as occlusion, moving objects or even software bugs.  Including these transient feature points in registration not only increases the amount of computation that is required, but also makes the algorithm more prone to errors.  As such, our main idea is to filter out these transient feature points, and retain the persistent ones as much as possible. The filtering algorithm using \metric is called \core.

In our implementation, we integrate \core into the scan-to-map LO framework. We keep only persistent features in the local feature map, while filtering out transient ones after a period of observation. \core can not only greatly reduce the feature point count in the local map as showed in Fig.~\ref{fig:show}, but also considerably reduce the constraints in the registration process. In the evaluation, we test two popular feature extraction algorithms with \core, and both prove that our algorithm can improve both the efficiency and the accuracy of LO. 

In summary, our main contributions are as follows:
\begin{itemize}
    \item We propose a new attribute \metric to evaluate feature points, which indicates whether a point is persistent or transient. Our \core algorithm carefully curates the suitable feature points by filtering out the transient ones;
    \item We integrate \core into the scan-to-map LO system, and conduct experiments on KITTI dataset\cite{kitti} to demonstrate that \core can improve the efficiency and accuracy at the same time.
    The  experimental  results  show  that  \core can remove about  48.4\%  of  the  points  in  the  local  feature  map  and  reduce feature points in scan by 19.3\% on average, which save 20.9\% time per frame on average and improves  the  accuracy  by  9.4\%.
\end{itemize}

%% file: chapters/relatedworks.tex
\section{Related Work}

\subsection{LiDAR Odometry}

LO has boomed in recent years. Since LOAM~\cite{zhang2014loam} in 2015, there have been various variants, like F-LOAM~\cite{floam}, LeGO-LOAM~\cite{legoloam}, LiLi-OM~\cite{liliom}, LOAM-Livox~\cite{livoxloam}, based on its framework. 
In particular,
LeGO-LOAM~\cite{legoloam} segments ground points and splits the optimization into two step to reduce search space. 
F-LOAM~\cite{floam} abandons the scan-to-scan match and replaces it by only scan-to-map with high frequency. 
Mulls~\cite{pan2021mulls} breaks down the types of features further and optimize the registration algorithm. 
LiLi-OM~\cite{liliom}  and LOAM-Livox~\cite{livoxloam} extent LOAM to solid-state LiDAR. 
\subsection{Feature Selection}

Feature selection considers how to extract subsets from raw data while maintaining  the  accuracy. 
ROI-cloud~\cite{zhou2020roi} presents a key region extraction method.
Aiming at removing dynamic and redundant point cloud, they voxelize 3D space into weighted cubes, and update the ``importance'' of each cubes continuously. Jiao et al.~\cite{greedy} use stochastic-greedy algorithm to select good features according to motion information on the premise of  preserving the spectral property of information matrices. 
Inspired by the Dilution Of Precision (DOP) concept in the
field of satellite positioning, KFS-LIO~\cite{li2021kfs} proposes a quantitative evaluation method for constraints derived from features and integrates the method in tightly-coupled LiDAR inertial odometry framework.
FasterGICP\cite{chenzonghai} incorporates acceptance-rejection sampling-based two-step point filter into the GICP method. Wan et al.~\cite{huangguoquan} formulate the uncertainty model, sensitivity model, and contribution theory to calculate the contribution of every residual term. With the idea similar to feature selection, Yin et al.~\cite{map_compression} scores the  points in map by observation count for map compression and Pomerleau, Fran{\c{c}}ois et al.\cite{longterm} computes whether a point
is dynamic or static based on multiple observations.

As a similar work, ROI-cloud~\cite{zhou2020roi} also updates status of points across consecutive frames.
However, ROI-cloud uses geometric information to evaluate the importance of each cube,  while we use \metric to evaluate \emph{persistence} level of a feature point.

\subsection{Semantic SLAM}

Using semantic information to understand the scene is also a way to extract persistent features. 
Dong et al.~\cite{pole} extract poles from point cloud to realize reliable and accurate localization. 
VI-Eye~\cite{vieye} exploits traffic domain knowledge by detecting a set of key semantic objects including road, lane lines. They extract a small number of salience points and leverage them to achieve accurate registration of two point clouds. This method achieves amazing experimental results, but relies heavily on semantic segmentation and spends the most time on it. 
In this paper, we intent to provide an approach to automatically extract the most informative regions by observing the historical registration process to achieve similar results without the help of semantic information.

%% file: chapters/preliminaries.tex
\section{Preliminaries}

Before presenting \core, we first review the formulation of the point cloud registration problem. Then, we introduce two methods to extract feature points which are widely used in LO. 
It has shown that these feature extraction methods can improve the efficiency and the robustness of the registration process.

\subsection{Formulation of Registration}
\label{sec:3.1}


 As specified in~\cite{survey_registration}, given two point clouds $\mathcal{P}_0$, $\mathcal{P}_1$ and their poses $\mathcal{X}_0$ and $\mathcal{X}_1$ in the same coordinate, the goal of registration is to estimate the transformation $\textbf{T}^*$ from $\mathcal{X}_0$ to $\mathcal{X}_1$. In the point-to-point ICP algorithm~\cite{icp}, $\textbf{T}^*$ can be calculated by:
\begin{equation}
\label{eq:registration}
\begin{aligned}
    \textbf{T}^* &= \mathop{\arg\min}_\textbf{T} \ \ \|d(\mathcal{P}_0, \textbf{T}(\mathcal{P}_1))\|^2 \\
\end{aligned}
\end{equation}
where $d(\cdot,\, \cdot)$ denotes the loss function to measure the distance, and 
$\textbf{T} =
\left[
\begin{matrix}
    \textbf{R} & \textbf{t}  \\
    \textbf{0} & 1 
\end{matrix}
\right]$, where $\textbf{R} \in SO(3)$ is the rotation matrix and $\textbf{t} \in \mathbb{R}^3$ is the translation vector.

\subsection{Feature Extraction}
\label{sec:3.2}

To reduce the number of considered points in ICP, feature extraction focuses on the local geometric structures of the point cloud. We review two of such methods that have been widely applied in LO.

The first feature extraction method applied in LOAM~\cite{zhang2014loam}, LeGO-LOAM~\cite{legoloam}, and F-LOAM~\cite{floam}, utilizes the unique format of point clouds of mechanical LiDAR. 
This method is also considered as the feature extraction method  based on ``ring''.
The $k$th frame $\mathcal{L}_k$ from LiDAR is composed of several parallel rings consists of points continuously. We denote $\textbf{p}^{(m,n)}_k$ for the $n$th point on the $m$th ring in the $k$th frame point cloud, then the smoothness $c^{(m,i)}_k$ of $\textbf{p}^{(m,i)}_k$ is calculated by the neighborhood of $\textbf{p}^{(m,i)}_k$ on the same ring.
More details can be found in \cite{zhang2014loam, legoloam, floam}. By setting some thresholds, two special types of points can be extracted from $\mathcal{L}_k$, called edge features, denoted as $\mathcal{E}_k$, and surface  features, denoted as $\mathcal{S}_k$. 

The second method applied in Mulls~\cite{pan2021mulls} and LiLi-OM~\cite{liliom}, is not limited to the special structure, i.e., the ring, of the point cloud from LiDAR. 
This method is also considered as the feature extraction method based on eigenvalues.
$\mathcal{L}_k$ is first stored in the KD-tree, and then the points denoted as $\mathcal{N}^i_k$ in the neighborhood of $\textbf{p}^i_k$ can be found in $O(log~n)$. $\mathcal{N}^i_k$ can be the nearest m points or the points within a sphere of radius~$r$. By calculating the eigenvalues $\lambda_1> \lambda_2>\lambda_3$ of the covariance matrix of $\mathcal{N}^i_k$, we can also define two variables to describe the local roughness of the point cloud. Linearity $\theta_l$ and planarity $\theta_p$ are defined as $\theta_l = \frac{\lambda_1 - \lambda_2}{\lambda_1}$ and $\theta_p = \frac{\lambda_2 - \lambda_3}{\lambda_1}$. Similarly, by setting some thresholds, the edge features and the surface  features can be extracted from $\emph{L}_k$.

With the edge features and the surface  features, the loss function $d(\cdot,\cdot)$ in Eq.~\eqref{eq:registration} can be replaced by the distance between edge features and edge extracted previously and the distance between surface  features and surface as follows:
\begin{equation}
\begin{aligned}
    d_e(\textbf{p}_e,\textbf{p}_e^M,\textbf{n}_e^M) &= \|(\textbf{T}\,\textbf{p}_e - \textbf{p}_e^M)\times \textbf{n}_e^M\|,\\
    d_s(\textbf{p}_s,\textbf{p}_s^M,\textbf{n}_s^M) &= (\textbf{T}\,\textbf{p}_s - \textbf{p}_s^M)\cdot \textbf{n}_s^M,
\end{aligned}
\label{eq:edgesurf}
\end{equation}
where $\textbf{p}_e$, $\textbf{p}_s$ are the edge and surface  features, $\textbf{p}_e^M$, $\textbf{p}_s^M$ are the geometric center of the corresponding edge and the corresponding surface and $\textbf{n}_e^M,\textbf{n}_s^M$ are the unit vector of the edge and the norm of the surface.

The two methods have different standards for the filtering of points and resulting different sets of extracted features. 
In our experiments, we also evaluate the performance of \core based on these different sets of extracted features, respectively.
More details will be specified in Sec.~\ref{sec5.4}.

%% file: chapters/method.tex
\section{Methodology}

\begin{figure}[t]
	\centering
	\includegraphics[width = \linewidth]{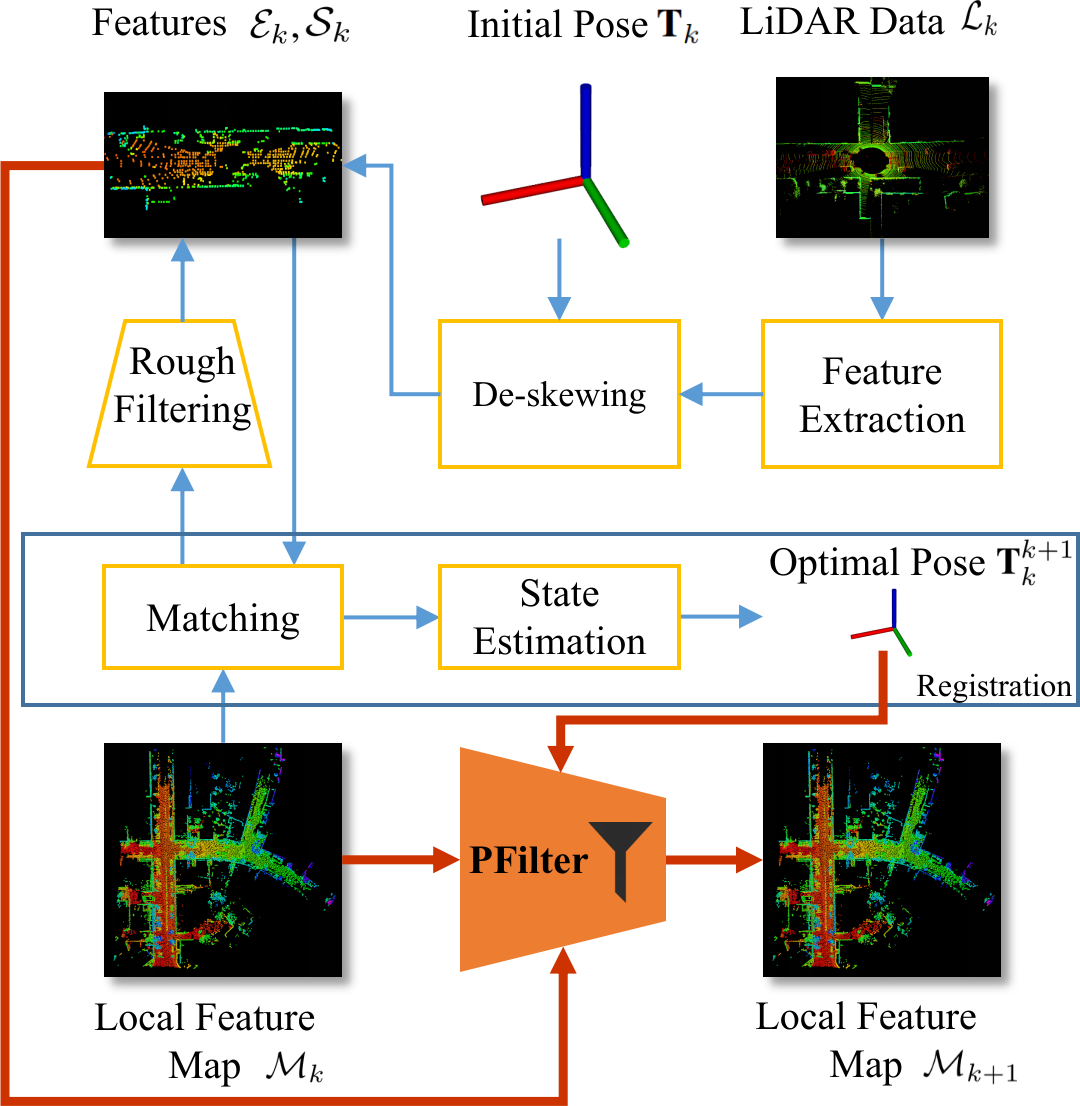}
	\caption{Pipeline of the scan-to-map LO with \core. The red arrows indicate the input and output of \core.}
	\label{fig:pipeline}
	\vspace{-2.0em}
\end{figure}

In this section, we first give the overview of our system.
Then we present the de-skewing and rough filtering in the system.
Finally, we discuss the definition of \metric and specify the details of \core.

\subsection{Overview}
\label{sec:4.1}

As shown in~\cite{cartographer,deschaud2018imls}, scan-to-map registration is generally more robust than scan-to-scan registration as the latter accumulates error faster. 
Then we use the scan-to-map LO (LiDAR odometry) system, F-LOAM~\cite{floam}, as the base for our discussion and evaluation. The two sources of registration in Eq.~\eqref{eq:registration} are feature points extracted from the scan and the local feature map, respectively.

As shown in the pipeline in Fig.~\ref{fig:pipeline}, for the $k$th frame, the input of the LO system consists of the three parts, i.e., a local feature map, denoted as $\mathcal{M}_k=\{\textbf{p}_0^M,\textbf{p}_1^M,\dots\textbf{p}_n^M\}$, a frame of point cloud generated by LiDAR (i.e., `LiDAR data' in the figure), denoted as $\mathcal{L}_k$, and an initial pose, denoted as $\textbf{T}_k$.
In particular, a local feature map $\mathcal{M}_k$ consists of two parts, i.e., $\mathcal{M}_k = \mathcal{M}_k^e\cup \mathcal{M}_k^s$, where $\mathcal{M}_k^e$ denotes the set of edge feature points and $\mathcal{M}_k^s$ denotes the set of surface feature points.
At the beginning, we apply the feature extraction method as discussed in Sec.~\ref{sec:3.2} on $\mathcal{L}_k$, which results in two sets of  feature points, i.e., edge features, $\mathcal{E}_k$, and surface features, $\mathcal{S}_k$. 
Later, these extracted features are further processed to reduce motion distortion, e.g., through de-skewing and  rough filtering that will be detailed in Sec.~\ref{sec:4.2}.
During registration, each point in the local feature map updates its \metric which measures the point's persistence level according to the match process, and will be definited in Sec.~\ref{sec:4.3}.
At the end, the local feature map will be filtered by \core according to the resulting features and the optimal  pose to remove transient features. The details of \core will be introduced in Sec.~\ref{sec:4.4}.

\subsection{De-skewing and rough filtering}
\label{sec:4.2}

To handle motion distortion, we assume that the LiDAR maintains the same motion in two adjacent frames, which is generally valid due to the high frequency of LiDAR (typically 10Hz). 
In specific, the distortion is corrected by applying a constant angular velocity and a linear velocity to predict the motion between two consecutive LiDAR scans, which generates better initial pose of each feature point for registration.

After assigning these initial poses for feature points, we apply rough filtering to remove incorrect matches between such a feature point and its corresponding points in the local feature map. The rough filtering is similar to the method used in LOAM~\cite{zhang2014loam}, which utilizes the distances between the feature point and its corresponding points, calculated by Eq.~\eqref{eq:edgesurf} and the geometric property of the corresponding points. After rough filtering, the feature points in qualified matches are used in registration for state estimation.

\subsection{Definition of \metric}
\label{sec:4.3}

In this section, we introduce the metric, \metric, to measure the persistence level of a feature point in the local feature map, which is weighed by how frequent the feature point has been detected in the recent frames.

In specific, we use $\textbf{p}_{k_0}^M$ to denote a feature point in the $k_0$th frame of the local feature map, i.e., $\textbf{p}_{k_0}^M\in \mathcal{M}_{k_0}$, where $k_0$ is the time when the point $\textbf{p}_{k_0}^M$ is first introduced in the local feature map. 
Then, for $k\geq k_0$, we define
\[
    \textbf{I}(\textbf{p}_{k_0}^M, k) = \left\{\begin{array}{ll}
    1 &  \exists\,\textbf{p}_k\in \mathcal{E}_k\cup \mathcal{S}_k, s.t.\  \textbf{p}_{k_0}^M\in \mathcal{N}_k(\textbf{p}_k),\\
    0 & \text{otherwise},
    \end{array}
    \right.
\]
where $\mathcal{E}_k$ and $\mathcal{S}_k$ denote the sets of extracted edge features and surface features from the $k$th frame of LiDAR data, respectively. 
$\mathcal{N}_k(\textbf{p}_k)$ denotes the set of corresponding points in $\mathcal{M}_k$ for $\textbf{p}_k$ from a resulting match after the rough filter process. 
Intuitively, $\textbf{I}(\textbf{p}_{k_0}^M, k)=1$ indicates that there exists an extracted feature point in the $k$th frame that can be matched with the point $\textbf{p}_{k_0}^M$.
In the following, a feature point $\textbf{p}_{k_0}^M$ is called to be {\em detected} in the $k$th frame, if $\textbf{I}(\textbf{p}_{k_0}^M, k)=1$.

Similar to the eligibility traces approach~\cite{rl}, we use \metric to specify how frequency and recency of a feature point that can be detected in past consecutive frames.
In particle, \metric of a feature point $\textbf{p}_{k_0}^M$ in the $k$th frame ($k\geq k_0$) is defined as:
\begin{align*}
\text{\metric}(\textbf{p}_{k_0}^M, k_0) &= \gamma\, \textbf{I}(\textbf{p}_{k_0}^M, k_0),\\
\text{\metric}(\textbf{p}_{k_0}^M, k) &= \gamma\, (\text{\metric}(\textbf{p}_{k_0}^M, k-1) + \textbf{I}(\textbf{p}_{k_0}^M, k)),
\end{align*}
for some $\gamma\in[0, 1]$.
Note that, the above recursive equations are algebraically equivalent to the following formula:
\begin{equation}
    \text{\metric}(\textbf{p}_{k_0}^M, k)=\sum_{\tau=k_0}^{k}\gamma^{k+1-\tau}\textbf{I}(\textbf{p}_{k_0}^M, \tau).
    \label{equation:define}
\end{equation}

\begin{figure}[t]
	\centering
	\subfigure[The point cloud at the 100-th frame. The green dotted box indicates a sparse point cloud and the detail in the white dotted box is shown in the top right.] {\includegraphics[width = 0.48\linewidth]{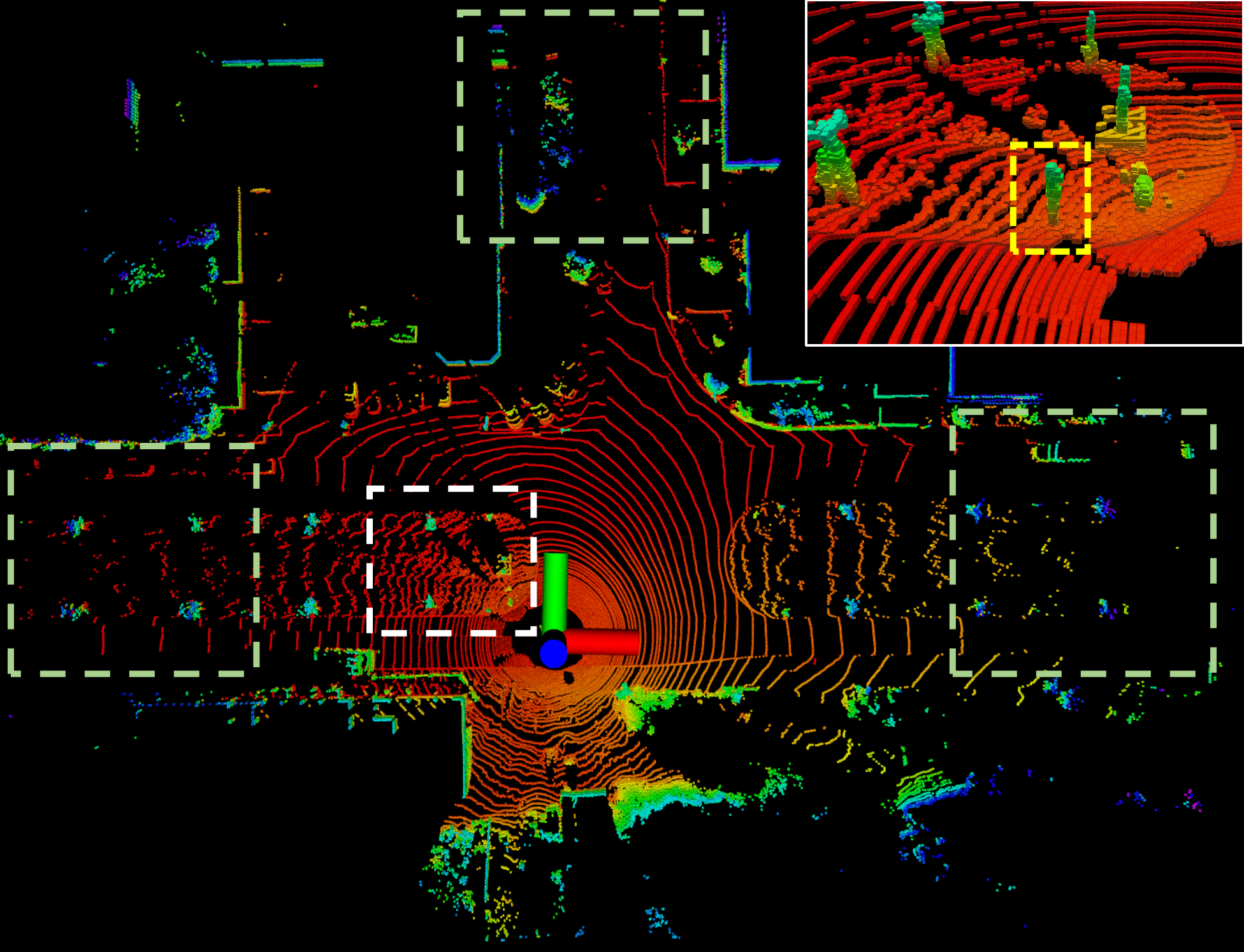}\label{fig:discount rate(a)}}
	\subfigure[The values of \metric w.r.t. consecutive frames for an edge feature point.] {\includegraphics[width = 0.48\linewidth]{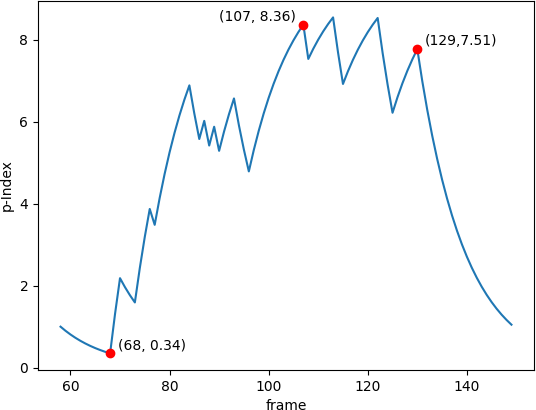}\label{fig:discount rate(b)}}\
	\vspace{-0.5em}
	\caption{Illustration of the \metric value. (b) illustrates the values of \metric for an an edge feature on a pole inside the yellow dotted box in (a). At the 58th frame, it first appears in the LiDAR's FOV. It is continuously detected from the 68th frame to the 107th frame. 
    Finally, only few (about three) points on the pole are scanned at the 129th frame; afterwards this edge feature is no longer detected.}
	\label{fig:discount rate}
	\vspace{-2.0em}
\end{figure}

Intuitively, a larger value of \metric is assigned to those feature points that have been detected more frequently recently.
In Eq.~\eqref{equation:define}, $\gamma$ specifies the effect of the time interval on the value of \metric. 
The reason for introducing $\gamma$ is that the point cloud is densely distributed nearby the agent and sparsely distributed in the distance, as illustrated in Fig.~\ref{fig:discount rate}. 
When the agent is moving forward, $k$ increases and landmarks near the agent at the $\tau$th ($\tau \leq k$) frame will be left behind. Then the point cloud at the $k$th frame around the landmark would be distributed sparsely, where the corresponding feature points would no longer be important. 
In particular,  when $k+1 -\tau$ increases, $\gamma^{k+1-\tau}$ would converge to $0$,  which reduces the weight of $\textbf{I}(\textbf{p}_{k_0}^M, \tau)$ in $\text{\metric}(\textbf{p}_{k_0}^M, k)$.

We set $\theta_{p}$ as the threshold to identify persistent feature points.
In specific, a feature point $\textbf{p}_{k_0}^M$ is {\em persistent} at the $k$th frame, if $\text{\metric}(\textbf{p}_{k_0}^M, k) > \theta_p$. Otherwise, $\textbf{p}_{k_0}^M$ is {\em transient}.
Note that, the larger the $\theta_p$, the fewer the persistent feature points. 

\input{chapters/alg}

\subsection{{Building Persistent Maps through \em \core}}
\label{sec:4.4}

Now we specify details of our feature filtering algorithm, \core. 
In specific, at the $k$th frame, we first update $\text{\metric}(\textbf{p}_{k_0}^M, k)$ for each feature point $\textbf{p}_{k_0}^M$ in the local feature map.
We also estimate an initial value of $\text{\metric}(\textbf{p}_k, k)$ for each feature point $\textbf{p}_k\in \mathcal{E}_k \cup \mathcal{S}_k$, i.e., extracted feature points from LiDAR data, using the average value of \metric for five feature points that correspond to $\textbf{p}_k$ in the match. 
Then we add these extracted feature points with their initial values of \metric to the local feature map. 
For each feature point $\textbf{p}_{k_0}^M$ in the local feature map,
if $\textbf{p}_{k_0}^M$ has just been detected recently, i.e., $k - k_0 < \kappa_{new}$ for a small natural number $\kappa_{new}$, then we maintain it in the local feature map for a while to see whether it would be persistent or not. 
If \metric of $\textbf{p}_{k_0}^M$ is larger than a constant $\theta_{max}$, i.e.,  $\text{\metric}(\textbf{p}_{k_0}^M, k)> \theta_{max} \in [\gamma, \frac{\gamma}{1-\gamma})$, then this feature point will be maintained in the local feature map permanently. 
At last, we remove the feature point from the  local feature map, if $\text{\metric}(\textbf{p}_{k_0}^M, k) \leq \theta_p$.

The process of \core is summarized in Alg.~\ref{alg1}.
\begin{itemize}
    \item The input involves a local edge map, denoted as $\mathcal{M}_k^e$, a local surface map, denoted as $\mathcal{M}_k^s$, the extracted edge and surface features, denoted as $\mathcal{E}_k$ and $\mathcal{S}_k$, and the optimal pose,  denoted as $\textbf{T}_k^{k+1}$.
    \item Lines 1-7: The loop intends to estimate the value of \metric for each extracted feature point in $\mathcal{E}_k$ (resp. $\mathcal{S}_k$).
    Meanwhile, \metric for each feature point in the local feature map is also updated in Line 4-6.
    \item Line 9: Extracted feature points with their initial values of \metric are introduced to the local feature map. 
    \item Lines 10-22: The loop intends to remove feature points from the local feature map, if their values of \metric are lower than $\theta_p$.
    Meanwhile, we reserve points that are recently detected or whose values of \metric are used to be larger than $\theta_{max}$. For the transient feature points, we remove them in line 21.
\end{itemize}

Notice that, \core can significantly reduce the number of feature points in the local feature map.
This can also reduce the computational cost for identifying legal matches in the rough filter process, as there would be fewer corresponding points in the local feature map that need to be considered for an extracted feature point from LiDAR data.

\input{tables/table1}
\input{tables/table2}

%% file: chapters/alg.tex
\begin{algorithm}[t]
\caption{\core}\label{alg1}
\KwData{$\mathcal{M}_k^e,\ \mathcal{M}_k^s,\ \mathcal{E}_k,\ \mathcal{S}_k,\ \textbf{T}_k^{k+1}$}
\KwResult{$\mathcal{M}_{k+1}^e,\ \mathcal{M}_{k+1}^s$}
\For{$\textbf{e}_k$ \emph{\textbf{in}} $\mathcal{E}_k$}{  
    $\textbf{e}_k \leftarrow \textbf{T}_k^{k+1}\textbf{e}_k$\;
    $\mathcal{N}_k(\textbf{e}_k) \leftarrow$ Find\_correspondence($\textbf{e}_k, \mathcal{M}_k^e,5$)\;
    
    \For{$\textbf{p}_{k_0}$ \textbf{in} $\mathcal{N}_k(\textbf{e}_k)$}{
        \text{\metric}$(\textbf{p}_{k_0},k)\leftarrow$\text{\metric}$(\textbf{p}_{k_0},k) + 1$\;
    }
    
    $\text{\metric}(\textbf{e}_k,k)$ = Average($\text{\metric}(\textbf{p}_{k_0},k)$ 
    \textbf{for} $\textbf{p}_{k_0}$ \textbf{in} $\mathcal{N}_k(\textbf{e}_k))$\;
}
$\mathcal{M}_{k+1}^e\leftarrow \mathcal{M}_{k}^e \cup \mathcal{E}_k$\;
\For{$\textbf{p}_{k_0}$ \emph{\textbf{in}} $\mathcal{M}_{k+1}^e$}{
    \uIf{\emph{\metric}$(\textbf{p}_{k_0},k)> \theta_p$}{
        \If{\emph{\metric}$(\textbf{p}_{k_0},k)\geq\theta_{max}$}{
            \text{\metric}$(\textbf{p}_{k_0},k)\leftarrow +\infty$
        }
        \text{\metric}$(\textbf{p}_{k_0},k+1)\leftarrow \gamma \cdot $\text{\metric}$(\textbf{p}_{k_0},k)$\;
        continue\;    
    }
    \ElseIf{$k - k_0< \kappa_{new}$}{
        \text{\metric}$(\textbf{p}_{k_0},k+1)\leftarrow \gamma \cdot $\text{\metric}$(\textbf{p}_{k_0},k)$\;
        continue\;
    }
    Delete($\textbf{p}_{k_0}$)\;
}

Compute $\mathcal{M}_{k+1}^s$ w.r.t. $\mathcal{S}_k$, $\mathcal{M}_k^s$ following the similar procedure.

\end{algorithm}

%% file: tables/table1.tex
\begin{table*}[!htp]
\center
\begin{threeparttable}
\vspace{0em}
\caption{ATE(\%) of LO systems on KITTI dataset}

\vspace{-1.0em}
\begin{tabular}{ccccccccccccc}
\label{table1}
LO Systems    & 00 & 01  & 02     & 03     & 04      & 05     & 06      & 07     & 08     & 09     & 10 & Mean      \\ \hline
LOAM\cite{zhang2014loam}  & 0.78 & {\color[HTML]{FE0000} \textbf{1.43}} & {\color[HTML]{3166FF} \textbf{0.92}} & {\color[HTML]{FE0000} \textbf{0.86}} & 0.71&0.57 & 0.65 & 0.63 & 1.12 & 0.77 & {\color[HTML]{FE0000} \textbf{0.79}}&{\color[HTML]{3166FF} \textbf{0.84}}\\

A-LOAM   & {\color[HTML]{3166FF} \textbf{0.69}} &  1.96 & 4.53 & 0.94 & 0.72 &  0.49 & 0.59& 0.42 & 1.04 & 0.73 & 1.00&1.19\\

LeGO-LOAM\cite{legoloam}        &1.38&28.03&2.14&1.21&1.27&0.91&0.80&0.74&1.40&1.25&1.70&3.71\\

F-LOAM \cite{floam}  & 0.70 &  1.95 & 1.04 & 0.95 & {\color[HTML]{3166FF} \textbf{0.68}} &  {\color[HTML]{FE0000} \textbf{0.49}} & {\color[HTML]{3166FF} \textbf{0.54}}& {\color[HTML]{3166FF} \textbf{0.42}} & {\color[HTML]{3166FF} \textbf{0.93}}& {\color[HTML]{3166FF} \textbf{0.70}}& 1.02&0.85\\ \hline

F-LOAM with \core   
& {\color[HTML]{FE0000} \textbf{0.61}} & {\color[HTML]{3166FF} \textbf{1.77}} & {\color[HTML]{FE0000} \textbf{0.80}} & {\color[HTML]{3166FF} \textbf{0.89}} & {\color[HTML]{FE0000} \textbf{0.65}} & {\color[HTML]{3166FF} \textbf{0.51}} & {\color[HTML]{FE0000} \textbf{0.47}} & {\color[HTML]{FE0000} \textbf{0.38}} 
& {\color[HTML]{FE0000} \textbf{0.91}} & {\color[HTML]{FE0000} \textbf{0.62}} & {\color[HTML]{3166FF} \textbf{0.88}} & {\color[HTML]{FE0000} \textbf{0.77}}\\ 

\end{tabular}
\begin{tablenotes}
    \footnotesize
    \item \textcolor[RGB]{254,0,0}{\textbf{Red}} and \textcolor[RGB]{49,102,255}{\textbf{blue}} indicate the first and second best results respectively.
\end{tablenotes}
\end{threeparttable}
\vspace{-1.0em}
\end{table*}

%% file: tables/table2.tex
\begin{table*}[]

\caption{Compared with F-LOAM} \label{table2}
\center
\vspace{-1.5em}
\begin{tabular}{|c|ccc|ccc|ccc|}

\hline

& \multicolumn{3}{c|}{Number of feature points in local feature map} & \multicolumn{3}{c|}{Number of constrains}    & \multicolumn{3}{c|}{Time(ms)}\\ \hline
seq & F-LOAM    & F-LOAM w/ \core & $\Delta(\%)$ & F-LOAM  & F-LOAM w/ \core & $\Delta(\%)$ & F-LOAM & F-LOAM w/ \core &$\Delta(\%)$\\ \hline
00  & 90,465    &50,345 &44.3    & 2,758  & 2,340 &15.1    &  54.1    & 43.1  & 20.3    \\ \hline
01  & 84,775    &30,273 &64.3    & 4,127  & 2,538 &38.5    &  68.8    & 51.1  & 25.7    \\ \hline
02  & 72,660    &42,494 &41.5    & 2,946  & 2,481 &15.8    &  52.6    & 42.4  & 19.4    \\ \hline
03  & 91,788    &47,109 &48.7    & 3,927  & 3,083 &21.5    &  62.9    & 50.0  & 20.5    \\ \hline
04  & 78,773    &33,086 &58.0    & 3,749  & 2,957 &21.1    &  63.4    & 49.6  & 21.7    \\ \hline
05  & 90,940    &52,487 &42.3    & 3,056  & 2,557 &16.3    &  55.3    & 44.6  & 19.3    \\ \hline
06  & 90,241    &44,861 &50.3    & 4,596  & 3,657 &20.4    &  66.8    & 53.3  & 20.2    \\ \hline
07  & 90,062    &47,752 &47.0    & 2,770  & 2,378 &14.2    &  53.3    & 41.7  & 21.8    \\ \hline
08  & 109,777   &56,327 &48.7    & 3,325  & 2,635 &20.8    &  63.5    & 47.9  & 24.6    \\ \hline
09  & 86,714    &47,100 &45.7    & 3,384  & 2,757 &18.5    &  59.0    & 47.1  & 20.2    \\ \hline
10  & 74,428    &43,756 &41.2    & 2,673  & 2,396 &10.4    &  49.6    & 41.5  & 16.3    \\ \hline
Mean& 87,329    &45,053 &48.4    & 3,635  & 2,925 &19.3    &  59.0    & 46.6  & 20.9     \\ \hline
\end{tabular}
\vspace{-2.0em}
\end{table*}

%% file: chapters/experiment.tex
\section{Experiments}

We implement our feature filtering algorithm \core and integrate it into the well-established scan-to-map LiDAR odometry framework, F-LOAM. 
We first evaluate the performance of the improved LO system on the KITTI dataset.
The experimental results show that \core can improve both accuracy and efficiency of F-LOAM. 
Later, we show that persistent feature points usually correspond to static landmarks and transient feature points usually correspond to  moving  objects  or  noise in various typical environments. 
Then, we construct an ablation study to examine the effects of values of corresponding parameters, i.e., $\theta_p$, $\theta_{max}$, and $\kappa_{new}$. 
At last, we compare the performance of \core based on two feature extraction methods in Sec.~\ref{sec:3.2}, respectively. 
The result shows that, \core is not sensitive to specific feature extraction methods. 

All experiments are conducted on a laptop with an AMD Ryzen 7 4800H CPU and 16GB memory. The parameter $\gamma$ in Eq.~\eqref{equation:define} is set to be 0.6.

\begin{figure}[t]
	\centering
	\subfigure[The trajectories of moving pedestrians] {\includegraphics[width = 0.45\linewidth]{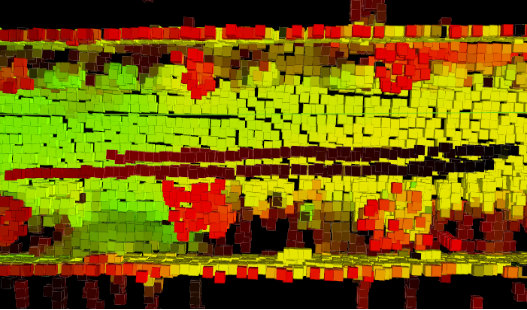}\label{fig:example:a}}
	\subfigure[The mistakenly extracted surface feature points] {\includegraphics[width = 0.45\linewidth]{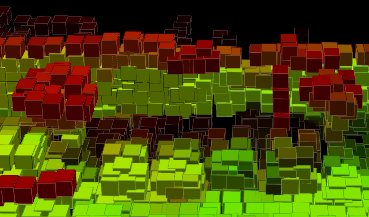}\label{fig:example:b}}
	\caption{The local surface feature map for an environment in KITTI.  The color of the feature points from green to red indicates the value of p-Index from high to low. The trajectories of moving pedestrians are indicated by the red-black lines in (a). (b) shows a detail of the environment involving the surface feature points correspond to static landmarks, but were extracted incorrectly. These transient feature points are identified by \metric automatically.
	}
	\label{fig:example}
    \vspace{-2.0em}
\end{figure}

\vspace{-0.5em}
\subsection{Performance on KITTI}
\label{sec5.1}

KITTI~\cite{kitti} odometry benchmark is one of the most popular datasets for  evaluating visual or LiDAR-based SLAM. 
The LiDAR used in the dataset is Velodyne HDL-64E, and the ground truth is given by the output of the GPS/IMU localization unit. It provides 11 sequences with ground truth that contains urban, country, and highway scenes. 
In our experiment, we set $\theta_p=1.5$, $\theta_{max}=2$, and $\kappa_{new} =2$.

Table.~\ref{table1} summarizes the accuracy of F-LOAM integrated with \core, as well as a few other LO systems. 
We use average translational error, ATE (m/100m), as in~\cite{kitti} to evaluate the accuracy of LO systems. 
We compare our improved F-LOAM with LOAM, A-LOAM, LeGo-LOAM, and the original version of F-LOAM. 

Table.~\ref{table2} summarizes the average number of feature points in local maps, the average number of constrains used to estimate ego-motion, and the average time spent on registration of F-LOAM with or without \core. 
We use $\Delta$ to denote the ratio of the reduction for the above metrics. Note that `w/' and `w/o' mean ``with'' and ``without'' in tables. 

As shown in Table.~\ref{table1} and~\ref{table2}, 
\core can significantly reduce the number of feature points for the registration process, can effectively alleviate the bottleneck, and can considerably improve the system accuracy. To be more precise, it removes about  48.4\%  of  the  points  in  the  local  feature  map  and  reduce feature points in scan by 19.3\% on average, which improves the efficiency of the SLAM system by 20.9\% and improve the accuracy by 9.4\%.



\subsection{Persistent Feature Points in a Real-world Environment}
\label{sec5.2}

As illustrated in Fig.~\ref{fig:example}, we show how \core works in a real-world environment. 
The environment in Fig.~\ref{fig:example} is a street in 
the KITTI 00 sequence with ground, wall, trees, and moving pedestrians. 
Fig.~\ref{fig:example} shows the surface feature map for the environment, where the color of the feature points from green to red indicates the value of \metric from high to low. Fig.~\ref{fig:example:a} shows the trajectories of moving  pedestrians which is composed of surface feature points extracted from the pedestrians at different frames. Fig.~\ref{fig:example:b} shows detail of the surface feature map. The surface feature points correspond to static landmarks including  crown of a tree, thin poles, and upper edges of walls are identified to be transient. These types of features with low \metric are filtered by \core automatically to reduce noise in the registration process.

\subsection{Ablation Study}
\label{sec5.3}

\begin{figure}[t]
	\centering
	\subfigure[\tiny{$\theta_p =2, \theta_{max}=\infty, \kappa_{new} =0$.}] {\includegraphics[width = 0.45\linewidth]{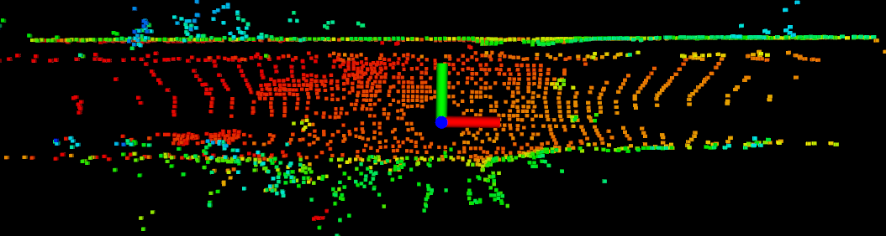}}
	\subfigure[\tiny{$\theta_p = 2, \theta_{max} =\infty, \kappa_{new}= 3$}] {\includegraphics[width = 0.45\linewidth]{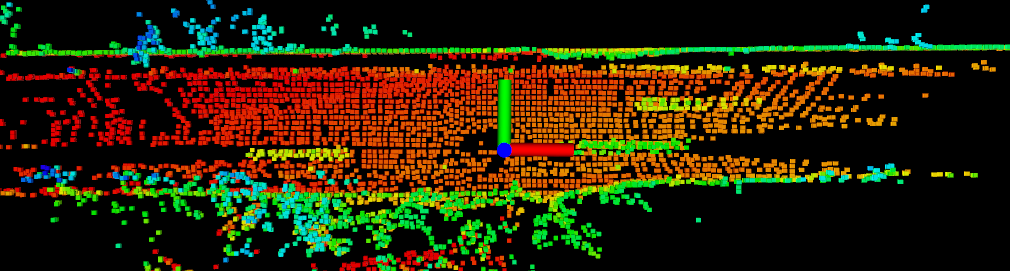}}
	\subfigure[\tiny{$\theta_p = 2, \theta_{max} =2, \kappa_{new} = 0$}] {\includegraphics[width = 0.45\linewidth]{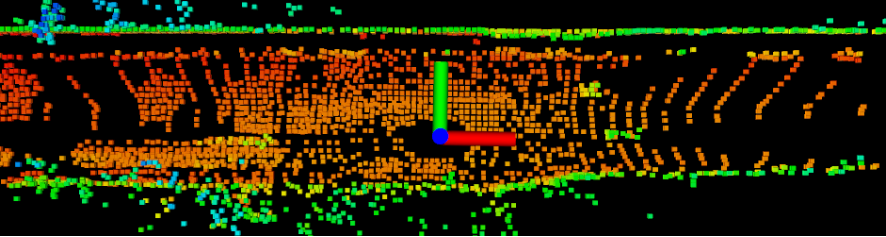}}
	\subfigure[\tiny{$\theta_p = 1, \theta_{max} =  \infty, \kappa_{new} = 0$}] {\includegraphics[width = 0.45\linewidth]{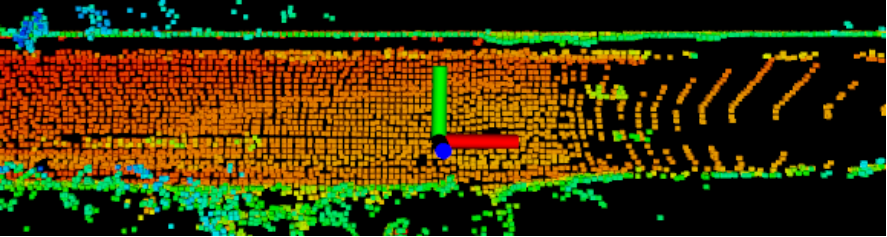}}
	\caption{The effects of $\theta_p, \theta_{max}$ and $\kappa_{new}$.}
	\label{fig:ablation_fig}
	\vspace{-2.0em}
\end{figure}


We conduct an ablation study to examine the effects of values of parameters in \core, i.e., $\theta_{p}$, $\theta_{max}$, $\kappa_{new}$.
We will examine the effect of one parameter by fixing other two parameters. 

As shown in Fig.~\ref{fig:ablation_fig}, $\theta_{p}$ is the threshold of \metric that mainly controls the number of persistent points. When $\theta_{p}$ increases, the number of persistent feature points decreases.
Note that, at the beginning, the noise is gradually filtered and the accuracy is improved.
However, when $\theta_{p}$ exceeds a certain threshold, the error starts to increase due to the lack of constraints.

$\theta_{max}$ and $\kappa_{new}$ are designed to improve the robustness of \core.
As shown in Fig.~\ref{fig:ablation_fig}, $\theta_{max}$ reserve more feature points corresponding to the landmarks behind the agent. $\kappa_{new}$ increase the number of feature points of where is new in FOV. Proper settings of $\theta_{p},\theta_{max}$ and $\kappa_{new}$ can improve the performance of the system.

\subsection{On Different Feature Extraction Methods}
\label{sec5.4}

We have introduced two method about feature extraction in Sec.~\ref{sec:3.2}, i.e., the one based on ``ring'' and the one based on eigenvalues.
F-LOAM applies the method based on ``ring''.
Of course, we can also replace it by the method based on eigenvalues and evaluate the performance of \core on the new system.
For a fair comparison, we consider the same number of feature points extracted by both methods. 
Table.~\ref{table_feature} summarizes the performance of the resulting system with or without \core. 
The  result  shows  that, \core is  not  sensitive  to  specific feature extraction methods.

\input{tables/table_feature}

%% file: tables/table_feature.tex

\begin{table*}[]
\caption{Performance of \core on Two Feature Extraction Methods}\label{table_feature}
\center
\vspace{0em}
\begin{tabular}{|c|cccc|cccc|}
\hline
\multirow{3}{*}{seq} & \multicolumn{4}{c|}{Based on ``ring''}                                                            & \multicolumn{4}{c|}{Based on eigenvalues}                                                                        \\ \cline{2-9} 
                     & \multicolumn{2}{c|}{Number of constrains}                         & \multicolumn{2}{c|}{ATE(\%)}          & \multicolumn{2}{c|}{Number of constrains}                                         & \multicolumn{2}{c|}{ATE(\%)}          \\ \cline{2-9} 
                     & \multicolumn{1}{c|}{w/o \core} & \multicolumn{1}{c|}{w/ \core} & \multicolumn{1}{c|}{w/o \core} & w/ \core & \multicolumn{1}{c|}{w/o \core} & \multicolumn{1}{c|}{w/ \core} & \multicolumn{1}{c|}{w/o \core} & w/ \core \\ \hline
00                   & \multicolumn{1}{c|}{2,758}  & \multicolumn{1}{c|}{2,340} & \multicolumn{1}{c|}{0.70} & 0.61 & \multicolumn{1}{c|}{2,487}      & \multicolumn{1}{c|}{2,087}                     & \multicolumn{1}{c|}{0.69}      & 0.61     \\ \hline
01                   & \multicolumn{1}{c|}{4,127}  & \multicolumn{1}{c|}{2,538} & \multicolumn{1}{c|}{1.95} & 1.77 & \multicolumn{1}{c|}{3,974}      & \multicolumn{1}{c|}{2,558}                     & \multicolumn{1}{c|}{2.26}      &2.06 \\ \hline
02                   & \multicolumn{1}{c|}{2,946}  & \multicolumn{1}{c|}{2,481} & \multicolumn{1}{c|}{1.04} & 0.80 & \multicolumn{1}{c|}{2,574}      & \multicolumn{1}{c|}{2,144}                     & \multicolumn{1}{c|}{0.98}      &0.83  \\ \hline
03                   & \multicolumn{1}{c|}{3,927}  & \multicolumn{1}{c|}{3,083} & \multicolumn{1}{c|}{0.95} & 0.89 & \multicolumn{1}{c|}{3,678}      & \multicolumn{1}{c|}{2,962}                     & \multicolumn{1}{c|}{0.80}      &0.71  \\ \hline
04                   & \multicolumn{1}{c|}{3,749}  & \multicolumn{1}{c|}{2,957} & \multicolumn{1}{c|}{0.68} & 0.65 & \multicolumn{1}{c|}{3,434}      & \multicolumn{1}{c|}{2,722}                     & \multicolumn{1}{c|}{0.79}      &0.78  \\ \hline
05                   & \multicolumn{1}{c|}{3,056}  & \multicolumn{1}{c|}{2,557} & \multicolumn{1}{c|}{0.49} & 0.51 & \multicolumn{1}{c|}{2,784}      & \multicolumn{1}{c|}{2,346}                     & \multicolumn{1}{c|}{0.52}      &0.53  \\ \hline
06                   & \multicolumn{1}{c|}{4,596}  & \multicolumn{1}{c|}{3,657} & \multicolumn{1}{c|}{0.54} & 0.47 & \multicolumn{1}{c|}{4,428}      & \multicolumn{1}{c|}{3,537}                     & \multicolumn{1}{c|}{0.54}      &0.51  \\ \hline
07                   & \multicolumn{1}{c|}{2,770}  & \multicolumn{1}{c|}{2,378} & \multicolumn{1}{c|}{0.42} & 0.38 & \multicolumn{1}{c|}{2,522}      & \multicolumn{1}{c|}{2,158}                     & \multicolumn{1}{c|}{0.49}      &0.37  \\ \hline
08                   & \multicolumn{1}{c|}{3,325}  & \multicolumn{1}{c|}{2,635} & \multicolumn{1}{c|}{0.93} & 0.91 & \multicolumn{1}{c|}{3,099}      & \multicolumn{1}{c|}{2,504}                     & \multicolumn{1}{c|}{0.96}      &0.95  \\ \hline
09                   & \multicolumn{1}{c|}{3,384}  & \multicolumn{1}{c|}{2,757} & \multicolumn{1}{c|}{0.70} & 0.62 & \multicolumn{1}{c|}{3,059}      & \multicolumn{1}{c|}{2,378}                     & \multicolumn{1}{c|}{0.63}      &0.75  \\ \hline
10                   & \multicolumn{1}{c|}{2,673}  & \multicolumn{1}{c|}{2,396} & \multicolumn{1}{c|}{1.02} & 0.88 & \multicolumn{1}{c|}{2,329}      & \multicolumn{1}{c|}{1,984}                     & \multicolumn{1}{c|}{1.10}      &1.14  \\ \hline
Mean                 & \multicolumn{1}{c|}{3,391}  & \multicolumn{1}{c|}{2,707} & \multicolumn{1}{c|}{0.85} & 0.77 & \multicolumn{1}{c|}{3,124}      & \multicolumn{1}{c|}{2,489}                     & \multicolumn{1}{c|}{0.88}      &0.84  \\ \hline

\end{tabular}
\vspace{-2.0em}
\end{table*}

%% file: chapters/conclusion.tex
\section{Conclusion}

In this paper, we consider how to improve the efficiency and accuracy of LiDAR odometry at the same time by carefully selecting feature points for registration. We observe that a large fraction of feature points are mistakenly extracted from individual LiDAR frames due to measurement errors or motions in the environment. Further, we find that these ``false'' feature points can be successfully identified by examining whether they continue to be detected from a sequence of frames in the near past. By filtering out those transient feature points, we can retain those persistent ones that correspond to stationary landmarks in the environment -- the true features that we should focus on in the registration. 

This is one of the earliest studies that explores the distribution of feature points across a stream of frames and exploits such distributions for more efficient and accurate SLAM. Moving forward, we will examine the feature points more closely to further cut down the false ones. We will also apply such techniques on other sensors such as cameras, radars, and different types of LiDAR sensors.   